\documentclass[conference]{IEEEtran}
\IEEEoverridecommandlockouts
\usepackage{cite}
\usepackage{amsmath,amssymb,amsfonts}
\usepackage{algorithmic}
\usepackage{graphicx}
\usepackage{lipsum}
\usepackage{textcomp}
\usepackage{xcolor}
\usepackage{hyperref}
\def\BibTeX{{\rm B\kern-.05em{\sc i\kern-.025em b}\kern-.08em
    T\kern-.1667em\lower.7ex\hbox{E}\kern-.125emX}}
\begin{document}

\makeatletter
\newcommand{\linebreakand}{%
  \end{@IEEEauthorhalign}
  \hfill\mbox{}\par
  \mbox{}\hfill\begin{@IEEEauthorhalign}
}
\makeatother

\title{Real-world adversarial attack on MTCNN face detection system}

\author{\IEEEauthorblockN{Edgar Kaziakhmedov\IEEEauthorrefmark{1},
Klim Kireev\IEEEauthorrefmark{2},
Grigorii Melnikov\IEEEauthorrefmark{3},
Mikhail Pautov\IEEEauthorrefmark{4},
Aleksandr Petiushko\IEEEauthorrefmark{5}}
\IEEEauthorblockA{
\IEEEauthorrefmark{1}\IEEEauthorrefmark{2}\IEEEauthorrefmark{3}\IEEEauthorrefmark{4}\textit{Skolkovo Institute of Science and Technology}; Moscow, Russia, \\
\IEEEauthorrefmark{1}\IEEEauthorrefmark{2}\IEEEauthorrefmark{3}\IEEEauthorrefmark{4}\IEEEauthorrefmark{5}\textit{Intelligent Systems Lab}; Huawei Moscow Research Center; Moscow, Russia\\
Email: \IEEEauthorrefmark{1}edgar.kaziakhmedov@skoltech.ru,
\IEEEauthorrefmark{2}klim.kireev@skoltech.ru,
\IEEEauthorrefmark{3}grigorii.melnikov@skoltech.ru,\\
\IEEEauthorrefmark{4}mikhail.pautov@skoltech.ru,
\IEEEauthorrefmark{5}petyushko.alexander1@huawei.com}}
\maketitle

\begin{abstract}
Recent studies proved that deep learning approaches achieve remarkable results on face detection task. On the other hand, the advances gave rise to a new problem associated with the security of the deep convolutional neural network models unveiling potential risks of DCNNs based applications. Even minor input changes in the digital domain can result in the network being fooled. It was shown then that some deep learning-based face detectors are prone to adversarial attacks not only in a digital domain but also in the real world. In the paper, we investigate the security of the well-known cascade CNN face detection system - MTCNN and introduce an easily reproducible and a robust way to attack it. We propose different face attributes printed on an ordinary white and black printer and attached either to the medical face mask or to the face directly. Our approach is capable of breaking the MTCNN detector in a real-world scenario.
\end{abstract}


\section{Introduction}
Contemporary deep learning systems are proved to be almost perfect face detectors, which outperform human abilities in this area {\cite{retina}}. The number of applications in today's life increases tremendously due to this fact. They would replace humans in areas where their accuracy is the most beneficial, for example, security. So since their algorithm-driven decisions may have serious consequences, the question of reliability and robustness against malicious actions becomes crucial. One of this task is face detection which is widely used as preparations operation for FaceID, which allows tracing criminals or control entrance policy. 

There are several deep learning approaches to this problem, end-to-end solutions like RetinaNet \cite{retina}, and cascaded from several NNs like MTCNN \cite{MTCNN}. Although end-to-end approach shows better results on synthetic benchmarks, cascaded systems with comparable quality are usually significantly faster. 

Unfortunately, there is a technique called adversarial attack, which allows deceiving almost any neural network-based systems in some instances. For example, in the case of a white-box attack in the digital domain, white-box since the attacker has access to topology and weights on the network, and digital domain because he changes input image pixel-wise. There is no existing solution to mitigate this issue according to recent publications {\cite{defences}} completely. Although these results are interesting from theoretical point of view, in practice, the task of face detection assumes that the processing image is obtained from a real-world device like camera, which is protected from the intrusion, i. e. attacker does not have direct access to the input. It is called physical domain attack. Although there are examples of this type of attacks, they proved to be much harder to perform, since adversarial attacks tend to be very fragile. Insignificant change in environment or illumination usually destroys them. In order to address this issue, a special technique called Expectation-over-Transformation (EoT) was introduced in \cite{EoT}. 

In this article, we present the attack on MTCNN face detection system. There are no published attacks on this face detector, although this system is quite well-known and public. Probably, the reason is that this system is robust to adversarial attacks due to its cascaded nature. Since it is hard to use traditional methods (FGSM-like) on the whole system, we decided to attack its first component. It is also worth mentioning that the attacking method implies the use of a public and well-known technique - adversarial attack; the network is available on the Internet and considered to be open, so the work does not violate any law or regulation.

The article is organised as follows. The attack itself is described in section III, the experiments in section IV, and in section II we review related works.

The source code and the video demonstration are available on the Internet \footnote{\url{https://github.com/edosedgar/mtcnnattack}}.

\section{Related works}

Before describing the proposed method, we review some of the widely used face detection models and their main differences. Then we focus on the adversarial attacks and consider some of the essential works in the area. Since most of the adversarial attacks are applicable in the digital domain and do not pose potential security concerns in applications using face detection models, we take a closer look at the real-world adversarial attacks and how they can be generated.

\subsection{Face detection models}

The problem of face detection was first practically solved in a seminal work of Viola and Jones 
\cite{violajones}. The idea was to apply a hand-crafted Haar feature to the input image pyramid of different scales. Multi-scale pyramid of images allows objects to be detected at different sizes and helps to narrow down the number of proposed regions, thus boosting up the classification. The algorithm is relatively fast, which enables real-time detection but at the cost of poor results with non-frontal faces and low light conditions. The following works \cite{improve1}, \cite{improve2}, \cite{improve3} mostly focused on improving the proposed architecture. 

Recent years have shown that CNN can potentially outperform all classic approaches based on standard features due to its generalization ability.  The approach that CNN utilized \cite{cnn_features} was quite similar to a classic one: the window with learnt features was sliding over the pyramid of input images and the resulted data fed to a fully-connected layer. Another way of constructing a face detection system was proposed in \cite{pyramid_feat} where authors suggested using the inherent multi-scale, pyramidal structure of DCNN to build feature pyramid.

Apart from the classification mentioned above, face detection networks can be categorized into two classes: single-shot and multi-shot detectors. One of the most well-known examples of the single-shot detector is SSD network \cite{ssd}, which takes an image as an input and computes a feature map with bounding boxes for each class. A similar approach was utilized in \cite{yolo}. Multi-shot detector (usually two-stage) suggests using several stages. The stages usually include proposal step and refinement steps. One of the most well-known examples of the two-stage detector is R-CNN \cite{rcnn}. The network extracts region proposal with the selected search algorithm and then warps cropped proposals to a square. Features are calculated based on a sparse set of candidates, and the output is fed to a classifier. 

The networks with a pure cascade architecture held leading positions for a long time in WIDER FACE challenge \cite{wider}. One of the most widely-used was MTCNN detector \cite{MTCNN}, which performs both face detection and face landmarking. The network uses three sub-networks: P-Net, R-Net and O-Net. The first stage does the coarse face detection producing proposal regions. Then Non-Maximum Suppression algorithm reduces the number of overlapping boxed forming more certain regions, which are fed to R-Net. R-Net refines the selected proposals, and O-Net does the face landmarking. MTCNN is still proposed to be used in the state-of-the-art face recognition system described in \cite{arcface}. Moreover, it is utilized in the most popular public face recognition implementation of FaceNet \cite{facenet} available on GitHub \footnote{\url{https://github.com/davidsandberg/facenet}}.

Besides excellent performance, MTCNN is a promising network in terms of robustness to adversarial attacks. The first shallow P-Net has a receptive field size of 12x12, which makes the detection resistant to the small artefacts on a face.

\begin{figure*}
\includegraphics[width=\textwidth,height=7cm]{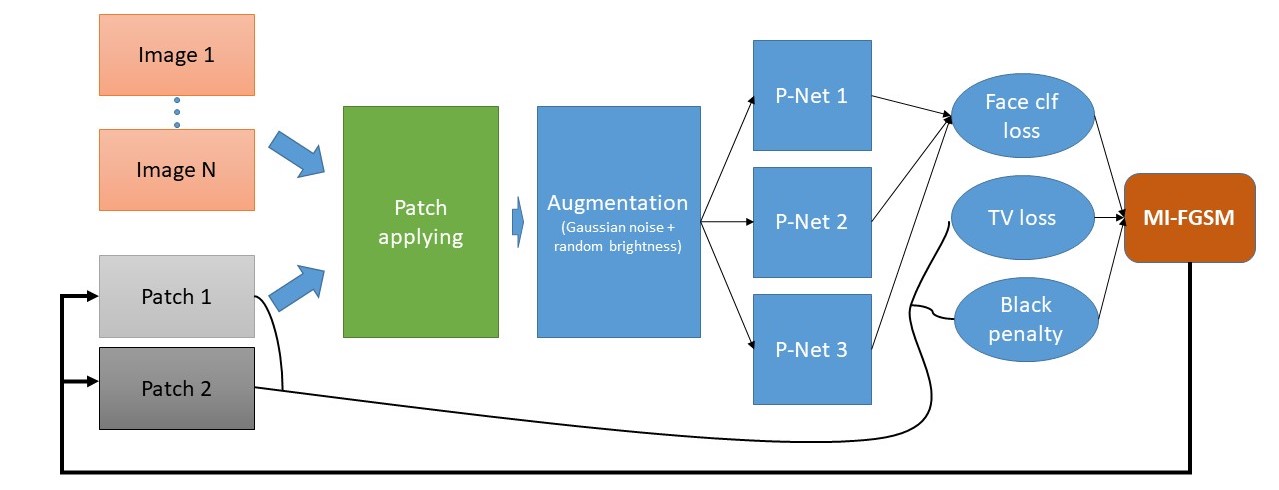}
    \caption{The attack pipeline. The patches are applied to the batch of images. The resulted images are augmented and fed to three networks. Loss of the face classification output and loss of the patches are computed, and the summed output is back-propagated to the patches.}
    \label{fig:architecture}
\end{figure*}

\subsection{Adversarial attacks}

Increased interest in neural networks helped to discover not only its state-of-the-art performance in various applications but also its serious drawback concerning robustness against adversarial attacks \cite{adv1}.
Minor and invisible to human eye changes in the input image can mislead the network and make it misclassify the object depicted in the picture with high confidence. It creates a severe security vulnerability in various applications of DCNN. 

Suppose a classifier mapping $f : \mathbb{R}^m \to \{1...k\} $ is given. The loss function is denoted by $L_f : \mathbb{R}^m \times \{1...k\} \to \mathbb{R}^+ $ and is to be continuous.
Provided $x \in \mathbb{R}^m$ is an image and $l \in \{1...k\}$ is a label, the problem can be formulated as follows:
\begin{align}
\min_{\begin{array}{c}\scriptstyle f(x+r) = l; \\[-4pt]
\scriptstyle x + r \in [0, 1]^m
\end{array}
} \label{eq} \| r \|
\end{align}

In \cite{adv2} it was proposed to use a box-constrained L-BFGS to find such $r$:
\begin{align}
\min_{x+r \in [0,1]^m} c\| r \| + L_f (x + r, l)
\end{align}

As soon as the method for finding adversarial examples was introduced, it was found in \cite{adv3} that neural networks have a linear nature. This fact enables attacks to be performed in a more computationally efficient way with the use of Fast Gradient Sign Method:
\begin{align}
    X^{adv} = X + \epsilon \text{sign} (\nabla_X J(X, y_{true}))
\end{align}

In some cases, one iteration of FGSM might not be sufficient; therefore, FGSM was extended to an iterative version in \cite{adv4}:
\begin{align}
    X^{adv}_{N+1} = \text{Clip}_{X, \epsilon} \Big\{ X^{adv}_N + \alpha \text{sign} (\nabla_X J(X_N^{adv}, y_{true})) \Big\}
\end{align}

In order to improve the optimization process further on and make the adversarial attacks more robust, momentum was added to iterative FGSM in the paper \cite{adv5} and the method is usually referred to MI-FGSM. It helps to memorize the gradient direction over iterations, which a right solution to pass through poor local minima or maxima.

It is commonly assumed, that network weights and model topology are known for an attacker.
That is called a white-box attack. It may occur that the model information is not available; thus FGSM cannot be applied; this is called a black-box attack, which is well-explained in \cite{adv6}.

\subsection{Attacks in physical domain}

The adversarial attacks in the physical domain are assumed to be more challenging. The input image fed to the network is subjected to various transforms imposed by the real-world: perspective transform, rotations, and so on. The searching procedure should take into account it and generate such an input that is tolerant of this kind of transforms. Such tolerance is usually achieved with a technique called Expectation over Transformation (EoT) \cite{EoT}. The key point of EoT is to model inherent perturbations in input during the optimization procedure.

Adversarial attacks in the real-world domain might impose more challenging problems as an attacker can mislead the network in a non-intrusive fashion. For instance, with the rise of autonomous vehicles, one can construct an attribute for a stop sign and fool a self-driving car \cite{carsign}. Authors did another illustrative work in \cite{yolofool}, where a particular pattern was generated to avoid detection by a person detector based on YOLO. The patch was trained with various transforms taken into account to enable the real-world attack.

\
In the article, we focus on a grayscale real face attributes to avoid face detection performed by MTCNN in real-time. The attack is supposed to be a white-box.

\section{Proposed method}

In this section, we describe the whole process of generating patches. Each sub-network of MTCNN has three output layers: face classification, bounding box regression and face landmark localization. Given this, we came up with three possible approaches for the attack:

\begin{itemize}
    \item Attacking the face classification layer of P-Net;
    \item Attacking the bounding boxes layer of P-Net, the similar method for YOLO was described in \cite{daedalus}, which exploits a non-secure NMS algorithm;
    \item Attacking the output layer of O-Net;
    \item Attacking the whole network.
\end{itemize}
The first approach compared to the others, requires the least demanding of architecture; hence, the face classification layer P-Net will be used for the attack. Refer to Figure~\ref{fig:architecture} for details of the proposed attack pipeline. In the following subsections, the detailed information on the attack pipeline will be given.

\begin{figure*}
\includegraphics[width=\textwidth,height=7cm]{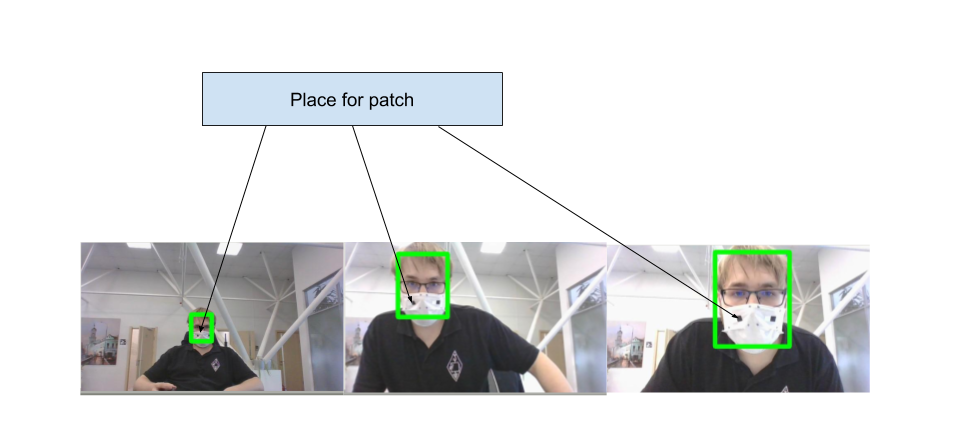}
    \caption{EoT process. The input images contain different head position and slightly different ambient luminance. The black dots on the mask depict placeholders for the patches.}
    \label{fig:eot}
\end{figure*}

\subsection{Expectation-over-Transformation}
For adversarial attack it is important to be robust to succeed in the physical domain. This task can be completed via EoT technique which was mentioned before. In our case we performed it in the following way: when an adversarial patch is being trained it does not minimize loss function over a single image, it uses batch consisted of multiple images with different positions of the head instead. Since it minimizes loss over pictures with different size of the patch and different brightness, it should be robust against these types of transforms in the real world. The scheme explaining the process is depicted in Figure~\ref{fig:eot}.

\subsection{Projective transform}

To apply rectangular patches on different surfaces, we use a projective map. Projective map is defined by its eight coefficients, which can be defined.
Firstly in the real world, we label patch location in edges of rectangles. If a patch has curved boundaries, it can be approximated with a rectangular grid. Then coefficients of projective transform are calculated, and the patch is applied. The example of how it is performed is depicted in Figure \ref{fig:prj}.

\begin{figure}
\centering
\includegraphics[height=4cm]{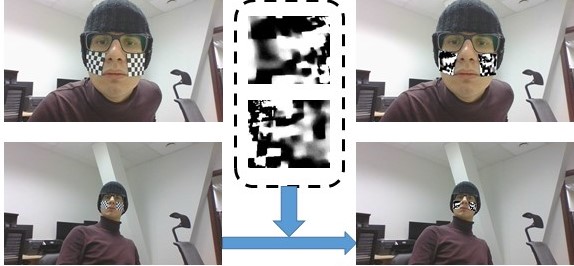}
    \caption{Example of a projective map from patches to the face. The face is labelled with printed grid, so for each rectangle coefficients of the transform can be defined.}
    \label{fig:prj}
\end{figure}

\begin{figure}
\centerline{\includegraphics[height=7cm]{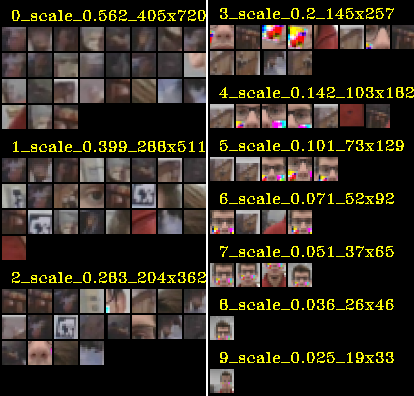}}
    \caption{The images fed to R-Net. The scales that give the most meaningful pictures for R-Net are 4, 5 and 7. These layers will highly contribute to the refinement stage.}
    \label{fig:analyzer}
\end{figure}

\subsection{MTCNN analyzer}
Once the patches are applied, and the resulted images are augmented, they should be resized to various scales and fed to P-Net. Originally, MTCNN builds up a pyramid of images with a given scale step factor. Using all scales for attacks is not feasible as it is more demanding of a resource. To mitigate this problem, we develop two possible approaches:
\begin{itemize}
    \item  We find the scale contributing most to the detection and use it with up-neighbouring scale and down-neighbouring scale;
    \item We find the scale that contributes most to the detection and use the scale with a slightly bigger size (which is not presented in a pyramid originally) and slightly smaller size, i.e. we do a size augmentation.
\end{itemize}

To find the most contributing scale, we let the image pass through the P-Net and manually trace the scale that gives the most meaningful results to R-Net. The example with pictures size of 24x24 is shown in Figure~\ref{fig:analyzer}. The more images with face passed to R-Net the more likely the face detection ends up successfully. Once three scales are selected in a way described above, the pyramid is created and loss functions of outputs are calculated.

\subsection{Loss functions}

The optimization process consisted of two main parts and one optional:

\begin{itemize}
    \item Face classification loss. The main objective is to lower the probability so that face will not be detected. Two losses were proposed to be used for $L_{clf}$: $L_{\inf}$ and $L_2$. Both showed comparably good results. As we use three layers, we sum the loss for each scale.
    \item Total variation loss. To make the optimization give preference for good-looking patterns without sharp color transitions and noise, we calculate $L_{TV}$ from patches given $p_{i,j}$ is a pixel value in position $i,j$:
    \begin{align}
    L_{TV} = \sqrt{(p_{i.j} - p_{i+1, j})^2 + (p_{i.j} - p_{i, j+1})^2}
    \end{align}
    Basically, the smoother transitions the less value of $L_{TV}$.
    \item Black penalty loss. In the case of surgical mask it is reasonable to reduce an amount of black color on the patch, that enables patch to be less unusual. To decrease the black colors, we use $L_{BLK}$:
    \begin{align}
    L_{BLK} = \sum_{i,j} 1 - p_{i,j}
    \end{align}
\end{itemize}

Finally, to balance the contribution of each optimization coefficients are added and the total loss has the following form:
\begin{align}
L = \sum_{i=1...3}{L_{clf_i}} + \alpha L_{TV} + \beta L_{BLK}
\end{align}

All coefficients were derived empirically and should be determined on a case-by-case basis.

\section{Experiments}

To do experiments, we need to define the main parameters for MTCNN. The minimal size parameter (minsize) determining the minimal size of an image in a pyramid was set to be 21 pixels (size of the shortest axis). The thresholds were set to 0.6, 0.7 and 0.7 for each sub-network respectively, and the scale step factor to 0.709. Such parameters are widely used in practice. 

The patches were trained for 2000 epochs; further training did not give any improvement. To test a generated pattern, we recorded videos with two setups: a surgical mask with patches and just two patches on cheeks. The videos were used to calculate a misdetection probability for various step scale factors.
To make the experiments more valuable in practical sense, the position in videos was different: close-up, mid-range and staying in a far distance. The results are given in Figure~\ref{fig:results}. It is worth mentioning that the performed attack was targeted, so not well-transferable to other persons not included to the training samples.

\begin{figure}
\centerline{\includegraphics[height=6.5cm]{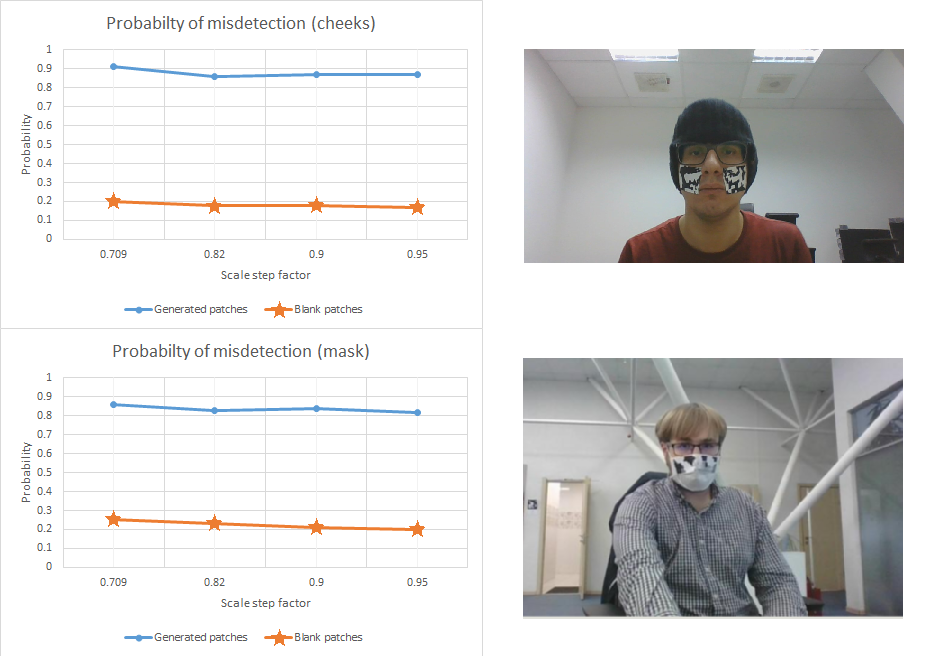}}
    \caption{Experiment results. The probability is averaged over 1000 frames for each scale step factor. With a generated patch, the misdetection probability becomes significantly higher in both cases. The photos on the right are two random frames from the recorded videos.}
    \label{fig:results}
\end{figure}

\section{Conclusion}
In this article we:
\begin{itemize}
    \item Found the method to attack the popular MTCNN face detector in the digital domain;
    \item Transferred this attack to the physical domain with EoT technique;
    \item Verified these results by conducting experiments in the real-world domain.
\end{itemize}

The obtained results show that one the most robust face detection network is still beyond expectations and need to be improved. The attacks in physical domain impose severe security issues, so possible ways to address it should be found.  In the article, we took the first step toward securing the face detection systems by finding a reproducible attack. In the future, we consider attacks using different places of intrusion in MTCNN detection pipeline and think about possible improvement of security.

\bibliographystyle{IEEEtran}  
\bibliography{ms}

\end{document}